# Learning of Art Style Using AI and Its Evaluation Based on Psychological Experiments

Mai Cong Hung[1,5], Ryohei Nakatsu[2], Naoko Tosa[3], Takashi Kusumi[4],
Koji Koyamada[2]

[1]Graduate School of Faculty of Science, Kyoto University, Japan
[2]Academic Center for Computing and Media Studies, Kyoto University, Japan
[3]Graduate School of Advanced Integrated Studies in Human Survivability, Kyoto University, Japan
[4]Graduate School of Education, Kyoto University, Japan
[5]RIKEN, Japan
hungmcuet@gmail.com, nakatsu.ryohei@gmail.com, tosa.naoko.5c@kyoto-u.ac.jp,
kusumi.takashi.7u@kyoto-u.ac.jp, koyamada.koji.3w@kyoto-u.ac.jp

**Abstract.** GANs (Generative adversarial networks) is a new AI technology that can perform deep learning with less training data and has the capability of achieving transformation between two image sets. Using GAN we have carried out a comparison between several art sets with different art style. We have prepared several image sets; a flower photo set (A), an art image set (B1) of Impressionism drawings, an art image set of abstract paintings (B2), an art image set of Chinese figurative paintings, (B3), and an art image set of abstract images (B4) created by Naoko Tosa, one of the authors. Transformation between set A to each of B was carried out using GAN and four image sets (B1, B2, B3, B4) was obtained. Using these four image sets we have carried out psychological experiment by asking subjects consisting of 23 students to fill in questionnaires. By analyzing the obtained questionnaires, we have found the followings. Abstract drawings and figurative drawings are clearly judged to be different. Figurative drawings in West and East were judged to be similar. Abstract images by Naoko Tosa were judged as similar to Western abstract images. These results show that AI could be used as an analysis tool to reveal differences between art genres.

## 1 Introduction

Recently, in AI, which mainly focused on logical processing, a method called "big data + deep learning" has emerged, and it has been found that this method is extremely effective in many areas. As the neural network used for deep learning resembles the structure of the human brain network, discussion on the possibility that AI simulates human intellectual ability and even exceeds it began to take place.

For Shogi (Japanese chess) and Go, AI software has recently surpassed the ability of human professional players. For example, Google-developed AlphaGo [1] defeated world top professional Lee Sedol in South Korea in 2016. Moreover in 2017, AlphaZezo [2] defeated AI chess, Shogi, and Go software, which were said to be the strongest at the time. In board games such as Shogi and Go, such capabilities as judging situation, using intuition, and searching deeply were considered to be the core of human intellectual capability and deeply related to human creativity. Therefore, it is now seriously discussed whether or not AI will surpass human intellectual ability, including human creativity [3].

Creativity is strongly connected to art. Artists create their artworks with their creativity. Then what kind of relationship can AI have with art? If AI has creativity, does the time come when AI becomes active even in the field of art creation and human artists are no longer needed? Such questions are becoming more and more real.

Recently a new technology of deep learning in AI called GANs (Generative Adversarial Networks) has been proposed [4], and as it becomes possible to perform deep learning with fewer training samples compared to conventional methods, various attempts to create artworks by AI have been carried out. However, many of these methods merely let AI learn the style of a particular painter and output images with the learned style. This means that so far AI does not have the ability to create artworks. Is there a different approach to the relationship between AI and art? For example, can AI approach basic questions such as what is beauty that exists at the basis of art, and what is the difference between Eastern and Western perceptions of beauty?

In this paper, firstly a topic of what art is and what beauty is will be discussed taking artworks of Naoko Tosa, one of the authors, as an example. Then a new methodology for approaching the relationship between AI and art will be proposed, and the results of verification through psychological experiments will be shown.

## 2    Related Works

### 2.1 Creation of art using AI

Until the "big data + deep learning" methodology [5] came out in AI training process, the structure of data feature was determined manually in advance. The great advantage of "big data + deep learning" is that once a big data is collected the network automatically analyzes the data structure and learns it by a multi-layer network without human preprocessing. With this method, a method of collecting and big data and training using it has become popular.

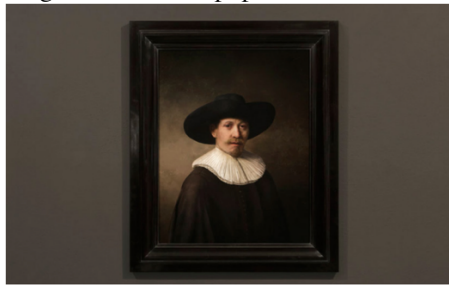

Fig. 1 Rembrandt-style portrait created by the "The Next Rembrandt" project.

Even in attempts to use AI in the field of art, a method of collecting a large amount of artwork as big data and letting AI be trained by it has been attempted. This allows

AI to learn artworks of a particular artist and to output an image with the artist's art style. A typical example of this is "The Next Rembrandt," a project run by a university and several companies [6]. In this project, first 346 paintings by Rembrandt were digitized and were used as data to train a multilayer network. As a result, Rembrandt's touch, color, layout, and other characteristics are stored as information in the multilayer network. Next, as the conditions such as a white man who looks at the right side, wears black clothes with a collar and a black hat is entered, the image that looks most like a painting by Rembrandt is obtained as an output from the multilayer network (Fig 1).

Recently, it has been announced that a painting produced by AI were sold by high price of 50 thousand dollars at one of art auctions [7]. The fact that a painting created by AI were traded at a high price in the art industry became a hot topic.

### 2.2 Learning of art style using AI

Deep learning uses a lot of layers that make up a neural network, so that the network learns the structure of given training data without having to perform pre-processing, feature data extraction, and other processing that was previously performed by humans. By doing so, it was revealed that AI has excellent identification capability. On the other hand, there is a problem that an extremely large amount of training data is required for training a multilayer network.

However, recently a new learning method called GANs (Generative Adversarial Networks), that can perform deep learning with a relatively small number of training data, has been proposed [4]. GANs are composed of two networks, a generator network and a discriminator network, as shown in Fig 2. For example, in the case of image generation/identification, the generator network learns to generate an image that cannot be identified by the discriminator network, while the discriminator network learns to perform more accurate identification. By performing learning as a zero-sum game between these two networks, deep learning can converge even with a relatively small number of learning data. By modifying this basic configuration, various GANs have been proposed and interesting results have been obtained.

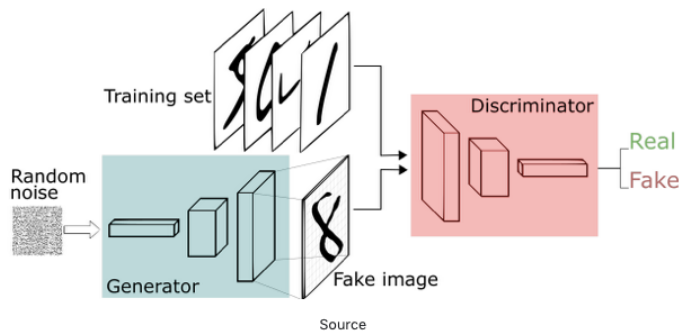

Fig. 2 Basic configuration of GANs.

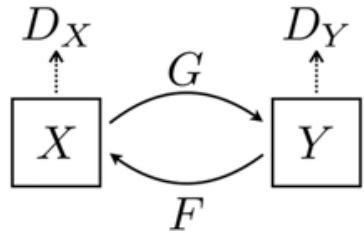

Fig. 3 Basic concept of Cycle GAN.

Among them, Cycle GAN is a new method that enables mutual conversion between two image sets. Fig. 3 shows the basic concept of Cycle GAN. In Cycle GAN, when two image sets (X, Y) are given, a transformation function G and an inverse transformation function F between them are considered. Also, two types of errors, Dx and Dy are considered; Dx is a difference between X and X' where X' is transformation of X by applying G then F and Dy is an error caused by the difference between Y and Y' where Y' is a transformation of Y by applying F and then G. The training is carried out so that the sum of these two types of errors is minimized.

Until then, GANs required a one-to-one correspondence between images belonging to two image sets. In Cycle GAN, however, even if there was no such correspondence between the conversion between image sets X and Y, the conversion between them is possible. By using this feature, for example, by learning a group of landscape photographs and a group of paintings of a specific painter (for example, Monet), mutual conversion between these image groups becomes possible. Figure 4 shows how Monet's paintings are converted into photographs using this feature, and Monet-style landscape paintings are created from landscape photographs [8].

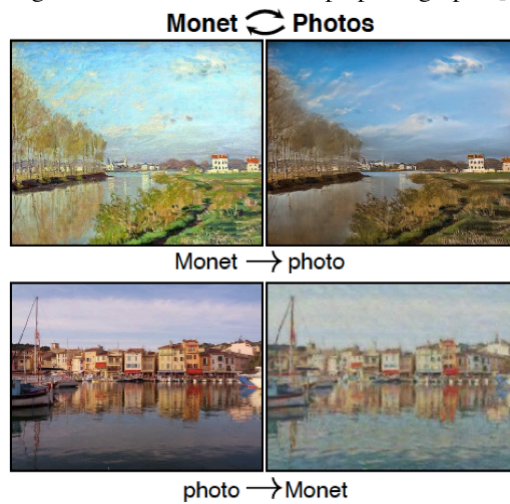

Fig. 4 Mutual conversion between landscape pictures and Monet paintings using Cycle GAN.
([8])

## 3. What is art

In this paper, a new framework of research to reveal the relationship between AI and art will be proposed in which by learning the styles of various artworks AI will clarify the characteristics of various art styles, and furthermore the essence of art. To that end, the art creation process called "fluid art" will be discussed and then what art and its underlying beauty is will be discussed.

### 3.1 Fluid dynamics and fluid art

The behavior of fluid is an important research subject in physics, and has been studied as "fluid dynamics [9]." It is known that fluid creates extremely beautiful shapes under various conditions. As beauty is a fundamental element of art, it is natural to consider fluid dynamics as a basic methodology of art creation. Naoko Tosa, one of the authors, has been leading a project of creating "fluid art" by shooting the behavior of fluid with a high-speed camera.

One good example of fluid based phenomenon is "milk crown." It is well known that when a drop of milk is dropped on milk and photographed with a high-speed camera, a beautiful shape like a crown is created. A video art called "Sound of Ikebana" inspired by this physical phenomenon is described in Section 3.3.

Another example of interesting behavior of fluid is "laminar flow" which is created when there is an obstacle in the fluid flow path. Another video art called "Genesis" was created inspired by this phenomenon [10].

### 3.3 Fluid art "Sound of Ikebana"

One of the techniques for creating fluid art is the creation of Ikebana-like shapes when sound vibration is applied to paint or other fluids and the phenomenon is shot with a high-speed camera. The detailed process is as follows. A speaker is placed upward, a thin rubber film is put on top of it, and a fluid such as paint is placed on top of it, and the speaker is vibrated with sound, then the paint jumps up and various shapes are created.

Naoko Tosa found that various fluid shapes could be generated by changing the sound shape, sound frequency, fluid type, fluid viscosity, etc. using this environment [11]. The resulting video image was edited to match the colors of the Japanese season, producing a video art called "Sound of Ikebana" [11]. Figure 5 shows one scene of the artwork. In April 2017, she exhibited Sound of Ikebana using more than 60 digital billboards at Times Square in New York. Figure 6 shows the event.

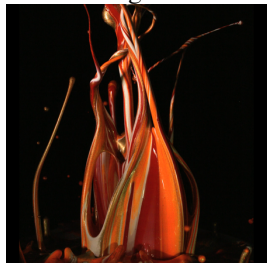

Fig. 5 Scene from "Sound of Ikebana"

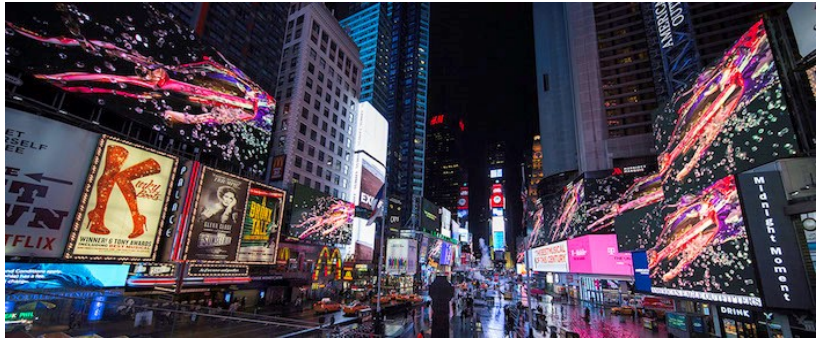

Fig. 6 "Sound of Ikebana" at Times Square in New York.

**3.4 What is beauty**

It was interesting that the video art mentioned above gave the authors the opportunity to consider what beauty is and also what Japanese beauty is. When Tosa, exhibited her media art around the world as Cultural Envoy named by the Agency for Cultural Affairs in Japan, many foreign art-related people indicated, "In Tosa's media art, which expressed the beauty hidden in physical phenomena, there is the beauty which Westerners did not notice until now, and which might be the condensed consciousness and sensitivity unique to Japan." After returning to Japan, she discussed with many Japanese art critics, curators, and researchers, and found that many agreed with this idea.

Based on this, the authors hypothesized that "Japanese beauty is the art of extracting the beauty hidden in nature [12]." How to verify this hypothesis is an important issue. The authors would like to study this through future research, but at the same time they are certain that AI could be used to study what is the characteristic of various art style and what is the beauty behind it.

## 4. Framework of This Research

As described in Chapter 2, Cycle GAN can be used to carry out transformation between two image sets, even if there is no one-to-one correspondence between images belonging to each image set. Figure 4 shows an example of conversion between a landscape photograph and a Monet landscape drawing. It can be seen that the mutual conversion has been successfully performed. This shows that AI can successfully learn the feature or style of Monet's paintings and put the styles on landscape photos (Fig. 7). So far, reference [8] merely states that landscape photographs could be converted into Monet-style paintings and vice versa. But the authors consider that Cycle GAN could be applied to a research investigating the relationship between art and beauty. This paper is based on this basic thought.

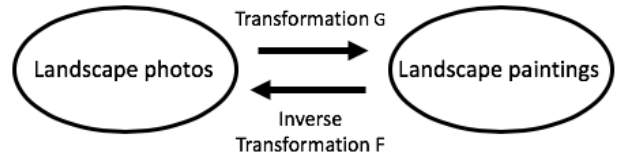

Fig. 7 Relationship between landscape images and landscape paintings viewed from AI

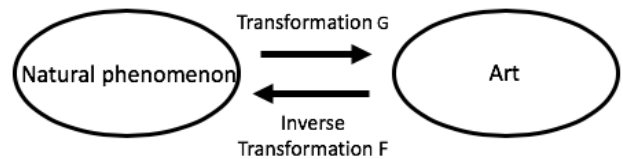

Fig. 8 Relationship between art, real objects, and natural phenomena viewed from AI

Figure 7 can be understood as Fig. 8 at abstracted level. Considering that art is an essential feature extracted from real objects or natural phenomena, it is possible to use Cycle GAN to convert between real objects or natural phenomena and art that extracts their essence.

There is a famous Aristotle's saying, "Art imitates nature [9]." This means that in the long history of art, painting originally tried to imitate nature. Western realism is the extension of this trend. As the times go down, however, the impressionism is born that tries to paint the light and its transitions perceived by human eyes, rather than trying to paint nature as it is. At this stage, the form of the objects depicted are still clear. Later, however, the history of Western painting was followed by cubism and surrealism, then followed by more recent abstract paintings. When it comes to abstract paintings, it is already at an unknown level what the paintings are expressing. But nevertheless, it can be said that artists extract essential things they felt in their hearts from the surrounding nature and made them into abstract paintings.

On the other hand, the history of Eastern painting is characterized by the fact that the objects painted have been clear since ancient times. Rather, it is characterized by the direction of minimalism that removes color like ink painting and by the way of drawing emphasizing the characteristics of the object like Ukiyo-e in Japan, and remains at the level of concrete painting compared to the West.

In such a situation, how is "Sound of Ikebana" described earlier positioned in the history of Eastern painting? It does not depict landscapes either human life, but at first glance is abstract images and videos. Nevertheless, as mentioned earlier, many people in overseas have said "The artwork has the feeling of Japanese beauty." Is it possible to use AI style learning and style conversion functions to find out how Sound of Ikebana is positioned compared to Western and Eastern concrete and abstract paintings?

In this study, this important and interesting issue is approached based on combining the function of art style learning and style conversion with AI and psychological experiments. Specifically, it was decided to proceed as follows.

(1) Two types of image sets (image set A, and multiple image sets B1, B2, ...) to be used to mutual conversion are prepared. The image set A is to be converted into art. For example, it could be photographs of landscape or a flower or fruit. A multiple image sets B1, B2, ... are composed of art images. For example, a set of

Impressionism paintings or a set of Cubism paintings. The images taken from artworks by Naoko Tosa described in Chapter 3 are also typical art images.

(2) Using Cycle GAN, mutual conversion (Fig. 9) of image set A and multiple image sets B1, B2, ... are achieved to obtain conversion functions (G1, F1), (G2, F2), …. If the conversion between the two image sets is successful, it indicates that there is a close association between these image sets. In other words, it can be said that the target art image is an extraction of the essential properties of the corresponding real image. Then how to evaluate whether the conversion was successful or not? A psychological experiment will be used for this.

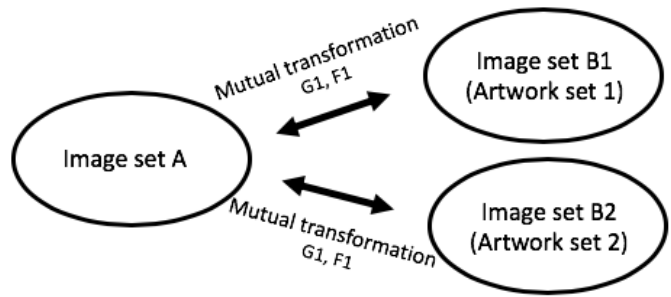

Fig. 9 Mutual conversion between two types of image sets.

(3) A psychological experiment is performed using the image sets G1 (A), G2 (A), ... that are obtained by performing the conversion G1, G2, … to the image set A. Depending on the purpose of the psychological experiment, questionnaires such as "Can you evaluate it as art?" "Do you feel beauty?" will be filled by the subjects of the experiment.

By doing this, it is possible to verify depending on different art styles what kind of information are extracted as essential information from real objects and made into artworks. For example, let the image set A be photographs of a real object such as flower. Let the artwork set B1 be the Tosa art described in Chapter 3 and the artwork set B2 be artworks of the Impressionism. In Chapter 3, the assumption that Tosa's artworks extract beauty from natural phenomena was made. On the other hand, it is thought that the artworks drawn by the Impressionists extract the beauty of flowers in the way of Western beauty. By asking subjects various questions about images G1(A) and G2(A), that were obtained by performing conversion using artwork set B1 and art work set B2, it is possible to clarify how Tosa's artwork is positioned in comparison with Western paintings and Eastern paintings.

## 5. Learning of Various Art Style and Transformation of Art Style

Based on the research framework described in Chapter 4, the following image sets were prepared.

Image set A: 8069 flower images
Image set B1: 1072 Monet art images of flowers
Image set B2: 123 Kandinsky art images
Image set B3: 238 Chinese hand-painted flower painting images called "Gongbi"
Image set B4: 569 images selected from "Sound of Ikebana"
(Resolution of all images are 256x256)

Image set A was prepared to perform mutual conversion with image sets B1, B2, B3, and B4 using Cycle GAN. The image set B1 is a painting mainly for flowers drawn by the Impressionist Monet as a representative example of the Western concrete paintings. As a representative example of Western abstract paintings, Kandinsky paintings were prepared as image set B2. Image set B3 includes flower paintings of Chinese hand-painted painting, called "Gongbi," as a representative example of Eastern concrete paintings. Image Set B4 is a set of still images taken from the media art "Sound of Ikebana" created by one of the authors, Naoko Tosa. The Sound of Ikebana is a video artwork created by shooting physical phenomenon with a high-speed camera. Basically, as it is made from physical phenomenon, it should not be said that it originally contains "Japanese beauty". But it was appreciated by people both inside and outside Japan as "including Japanese beauty." In this experiment, the art style of the Sound of Ikebana was compared with Western and Eastern representative painting styles.

The four types of different image sets B1, B2, B3, B4 and the image set A including were mutually converted using the Cycle GAN. Figure 10 shows examples of the result of the mutual conversion between the Sound of Ikebana and the photograph of flower.

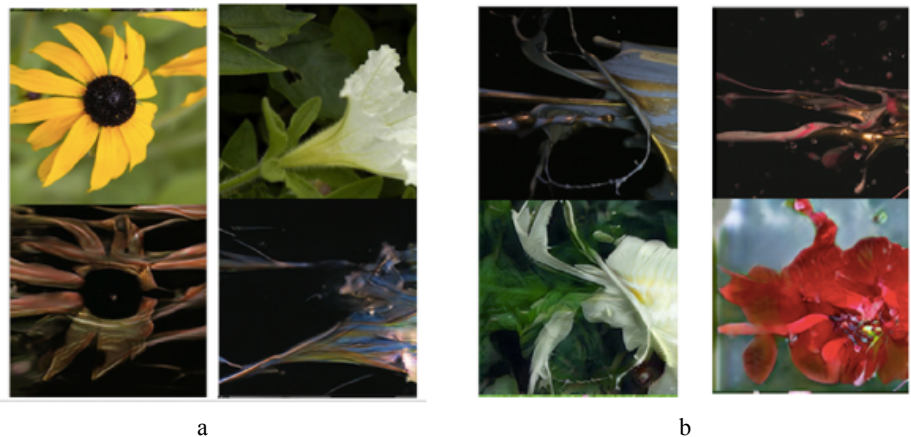

a  b

Fig. 10 Example of mutual conversion between flower photos and Sound of Ikebana style images. (a: Flower photos to "Sound of Ikebana" style images, b: "Sound of Ikebana" to flower-like images)

The question here is why it is necessary to use AI for comparative evaluation of art styles. Why Monet art, Kandinsky art, Chinese Gongbi art, and Sound of Ikebana images were not directly compared and evaluated using psychological experiments. There have been some studies that evaluated art works through psychological experiments [14] [15] [16] [17]. However, by using a copy of the original artwork, it can be relatively easy to identify the artist for each art. For example, knowing that a painting is Monet's art suggests that a subject has a prejudice of the work of Monet, a representative of the Impressionists in Western art history, and that this will have a significant effect on evaluation experiments. In order to avoid this effect, there are research examples of using lesser-known works [17]. However, works of famous artists and art schools can be easily identified and would greatly affect evaluation experiments. On the other hand, using GAN allows for AI to learn an art style and to apply the art style on the input images with the art style from the input. Therefore, bias can be avoided in the evaluation experiment and this is the benefit of using AI to evaluate artworks.

# 6. Evaluation of Obtained Results Based on Psychological Experiment

The experiment described in Chapter 5 yielded results of performing various style conversions on flower images. By having people evaluate the results of applying various style transformations to various flower images, is it possible to know what art is, what is the beauty behind it, and the culture of beauty? Is there any suggestion on how people receive Japanese beauty and the corresponding Western beauty? That is the goal of this research. Since this is a subjective evaluation, a method used in psychological experiments, which is to present a target image to a subject, to conduct a questionnaire survey, and sto tatistically analyze the results, was used.

### 6.1 Psychological experiment

Image group 1, 2, 3, and 4 are prepared by selecting two images from each of image sets G1(A), G2(A), G3(A), and G4(A), which are obtained by converting image set A into image sets B1, B2, B3, and B4. The resolution of each image is 256x256.

23 Kyoto University students (12 male and 11 female) were used. The gender ratio is almost half. Each of 8 images was printed out on A4 high-quality paper, and the eight images were presented to the subject who were asked to answer the prepared questionnaire. The order of the presented images was set randomly for each subject.

The subjects were asked to perform a seven-step subjective evaluation of the 21 items shown in Table 1. This excludes items that have small significant difference in the evaluation results due to differences in art types from the 23 adjective pairs used in [17], and added an item that directly asks whether it looks like Eastern or not, such as "Eastern-Western".

Table 1 Adjective pairs used for evaluation

| | |
|---|---|
| Individual - Ordinary | Complex - Simple |
| Masculine - Feminine | Soft - Hard |
| Dynamic - Static | Bold - Careful |
| Bright - Dark | Lively - Lonely |
| Warm - Cold | Sharp - Dull |
| Flashy - Sober | Like - Dislike |
| Deep - Superficial | Difficult to understand - Easy to understand |
| United - Disjointed | Beautiful - Ugly |
| Heavy - Light | Artistic - Non artistic |
| Vivid - Cloudy | Eastern - Western |
| Stable - Unstable | |

### 6.2 Analysis

The results of the subjective evaluations by 23 subjects were averaged for each evaluation item, graphed, and t-tested. Among the 21 items, the differences between each image group were relatively large for six items: Individual-Ordinary, Dynamic-Static, Stable-Unstable, Bold-Careful, Artistic-Non artistic, Eastern-Western. Figures

11 – 13 show the results of the average value and the standard error for each evaluation item. Also, the result of t-analysis (**:1%, *:5%) is shown on these figures.

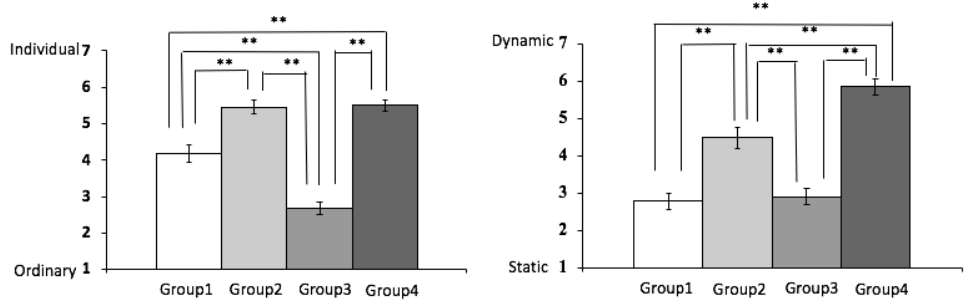

Fig. 11 Subjective evaluation results for "individual-ordinary" (left) and "dynamic-static" (right).

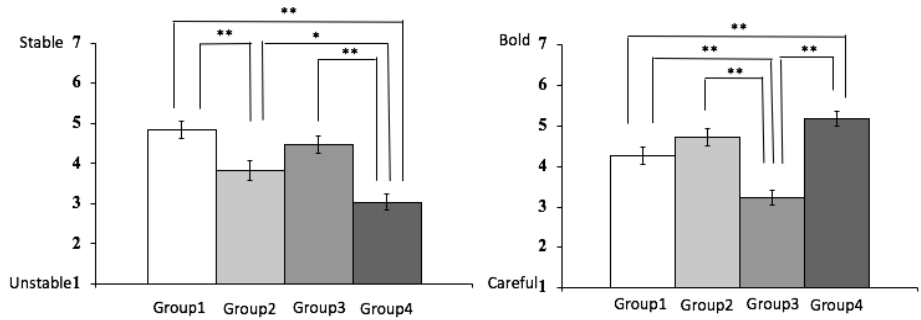

Fig. 12 Subjective evaluation results for "Stable-Unstable" (left) and "Bold-Careful" (right).

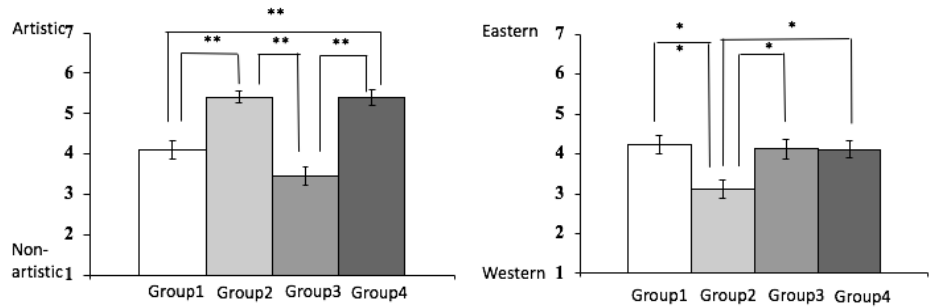

Fig. 13 Subjective evaluation results for "Artistic-Non-artistic" (left) and "Eastern-Western" (right).

**6.5 Consideration**

The following is the result of analysis from Figs. 11 to 13.

(1) Group 2 and Group 4

As can be seen from Figs 15 to 19, Group 2 and Group 4 received similar evaluations. In particular, the hypothesis that there is a significant difference between Groups 2 and 4 has been rejected in the items of "Individual-Ordinary", "Bold-Careful", and "Artistic-Non artistic." This indicates that there is little significant difference between the style of Kandinsky and the style of Sound of Ikebana. Conversely, Group 4 is evaluated as having a significant difference of 1% level from Group 1 and Group 3 for all items except "Eastern-Western". This indicates that the Sound of Ikebana is considered to be abstract rather than concrete. If the transition from figurative painting to abstract painting is the history of Western painting, the Sound of Ikebana can be said to be positioned in the history of the transition from figurative painting to abstract painting in the East.

(2) Group 1 and Group 3

Similarly, as can be seen from Figures 15 to 19, Group 1 and Group 3 receive similar evaluations. In particular, the hypothesis that there is a significant difference of 5% level between the "Dynamic-Static" and "Artistic-Non artistic" items have been rejected. Group 1 is an image set with a style of Western Impressionism, and Group 3 is an image set with the characteristics of an Eastern concrete painting. Each of the original art could be judged that each has its own characteristics, one on the Western side and the other on the Eastern side. But in a more essential part, these styles may have something in common.

(3) Artistic or not

It is interesting to note that Group 1 and Group 3 are around or below the median value of 4 for "Artistic-Non artistic." Few people will rate Monet's original image as unartistic. Chinese hand-painted paintings have also been highly evaluated as elaborately depicting nature. However, groups 2 and 4 are evaluated as having higher artistic than Groups 1 and 2. As described in [17], in the previous evaluations of artworks, concrete artworks were often evaluated higher than abstract artworks. But this psychological experiment shows a different tendency. This is thought to be due in part to the young age of the subject. As the younger generation has more opportunities to watch abstract drawings and recent media art, it can be said that they have come to have an aesthetic sense appreciating abstract paintings. Another cause is that instead of evaluating the original artwork itself, its style is extracted from the original image by AI, and the image with the style attached to photographs such as landscapes and still life. Therefore, it can be said that this is due to the characteristics of the basic procedure of this research. By using the original art image for evaluation, it was relatively easy to identify who the artist was, or even the specific work itself, and it is guessed that this would have a significant effect on the evaluation.

(4) Eastern or Western

As shown in Fig. 20, the answers to the question of Eastern or Western are all around the median of 4 except for Group 2. This indicates that the subject did not clearly answer the question of whether it was Eastern or Western, and gave a response near the middle. Initially, the authors expected that the Sound of Ikebana would be evaluated as "Eastern" because of the overseas evaluation that the Sound of Ikebana contains Japanese beauty, as described earlier. But so far, such result was not obtained.

But this does not mean that the Sound of Ikebana is not "Eastern". At the same time Monet-style images and Chinese Gongbi images were evaluated as intermediate. This seems to indicate that at this time, the art style extracted by AI has not yet reached a level to identify Western or Eastern impression. In the future research is necessary to consider how to raise the resolution of the image and make it a moving image so that it can be evaluated as a Western or Eastern style.

On the other hand, it is interesting that Group 2 is evaluated as Western. This can be understood from the fact that the subject found a typical abstract painting style in the obtained image. It shows that people have a common notion of "the abstract painting style." In other words, the appearance of abstract painting in the West has so much significance in art history.

## 7. Conclusion

In this paper, a new method of handling art with AI was described. With the emergence of a new deep learning method called GANs, that can carry out deep learning using relatively small training data, recently several attempts to handle art with AI have been made. However, most of these attempts are to make AI learn the artworks created by a specific artist or art school as training data, and to use the learned AI to create a work similar to the work of a specific artist or art school. Although it has been often claimed that AI can create art by this methodology, it is not a correct claim. These attempts have only produced works that are similar to those created by certain artists and art schools. Artists have been trying to create new methods one after another while making the most of their creativity. Such a creative process is not yet at the stage where the current AI can do. Rather than such an approach, there should be another approach by using GANs to investigate where is the difference in art style, and what is the essence of the difference in aesthetic sense between Eastern and Western beauty due to cultural differences. Such approach would be one step closer to the essential issues related to art.

By using the method proposed in this paper, subjects were able to evaluate the styles of Western and Eastern concrete and abstract paintings without bias created by identifying specific artists and/or artworks. As a result, regardless of the cultural differences between the East and the West, it was shown that the concrete drawings of the East and the West are not different. Also, it was shown that one of the authors' work "Sound of Ikebana," has no significant difference from the abstract drawings of Kandinsky. This means that the Sound of Ikebana gives the impression similar to the abstract picture of Kandinsky.

However, it is not enough to clarify why the Sound of Ikebana is evaluated by Westerners as Eastern in the scope of this study. It is a future work to clarify this, and we plan to conduct the following research.

(1) Improvement of image resolution

The resolution of the images used in this study is 256x256. Although as far as the image is printed out on A4 size paper, there is no particular problem in obtaining the evaluation, low resolution is noticeable when displaying them on a large screen. It is desirable that higher resolution such as HD resolution at least, and hopefully Full HD or even 4K resolution is realized. Increasing the resolution is one direction of future research.

(2) Evaluation of video art content

One of the reasons why the Sound of Ikebana is said to be Eastern is probably because of the slow movement of the video art of Sound of Flower. In order to verify this, it is necessary to create a moving image instead of a still image and evaluate it using a moving image.